\def\year{2022}\relax
\documentclass[letterpaper]{article} 
\usepackage{aaai22}  
\usepackage{times}  
\usepackage{helvet}  
\usepackage{courier}  
\usepackage[hyphens]{url}  
\usepackage{graphicx} 
\usepackage{natbib}  
\usepackage{caption} 
\DeclareCaptionStyle{ruled}{labelfont=normalfont,labelsep=colon,strut=off} 
\usepackage{algorithm}
\usepackage{algorithmic}
\usepackage{enumitem}
\setitemize{noitemsep,topsep=0pt,parsep=0pt,partopsep=0pt}
\usepackage{booktabs}
\usepackage{multirow}
\usepackage[normalem]{ulem}
\useunder{\uline}{\ul}{}
\usepackage{color}

\usepackage{comment}
\usepackage{mathrsfs}
\usepackage{amsmath}
\usepackage{amssymb,amsfonts}

 \usepackage[pagebackref=true,breaklinks=true,letterpaper=true,colorlinks,bookmarks=false,citecolor=cyan]{hyperref}

\usepackage{newfloat}
\usepackage{listings}

\floatstyle{ruled}
\newfloat{listing}{tb}{lst}{}
\floatname{listing}{Listing}
\title{Geometry-Contrastive Transformer for Generalized 3D Pose Transfer}
\author{Haoyu Chen$^1$, Hao Tang$^2$, Zitong Yu$^1$, Nicu Sebe$^3$, Guoying Zhao$^{1,}$\thanks{Corresponding Author.}
    \\
    $^1$CMVS, University of Oulu \quad 
    $^2$Computer Vision Lab, ETH Zurich \quad 
    $^3$DISI, University of Trento
    \\
    {\tt\small \{chen.haoyu, zitong.yu, guoying.zhao\}@oulu.fi} \\
    {\tt\small hao.tang@vision.ee.ethz.ch \quad nicu.sebe@unitn.it}
}

\begin{document}

\twocolumn[{%
\renewcommand\twocolumn[1][]{#1}%
\maketitle

\begin{center} \small
    \centering
    \includegraphics[width=1\linewidth]{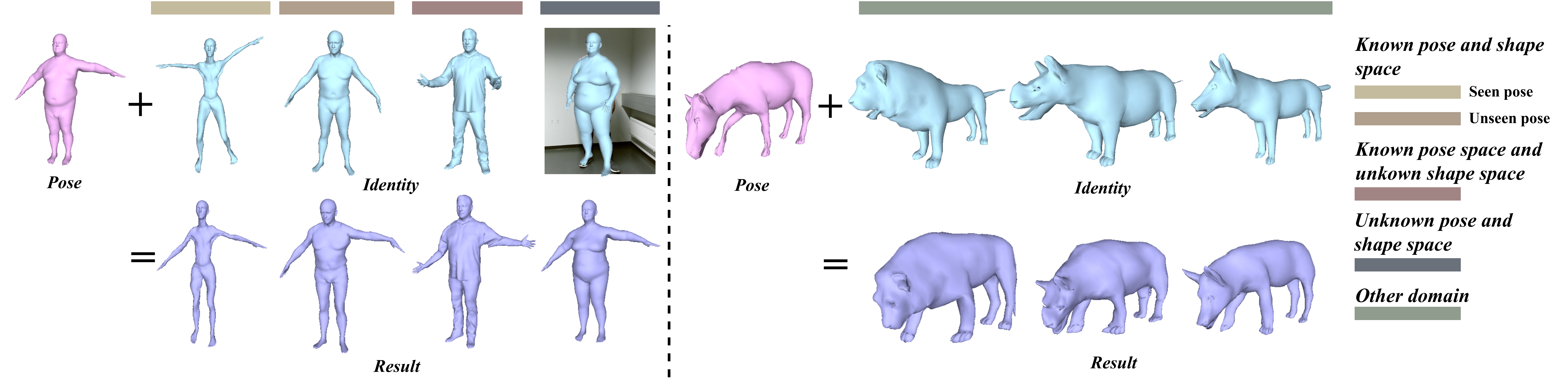}
    \captionof{figure}{Examples of pose transfer results by our 3D GC-Transformer. Blue, pink, and purple colors stand for identity, pose, and result meshes, respectively. The left part shows the human pose transfer results. The identity meshes are from FAUST \cite{FAUST}, MG-cloth \cite{MG-cloth}, SMPL-NPT \cite{NPT}, and our new SMG-3D dataset. The right part shows animal pose transfer results on the SMAL dataset \cite{SMAL}. Our method can be generalized to different spaces and even real-world scenarios and animals. More experimental results can be found in supplementary materials. \label{fig:1}}
\end{center}%
}]

\begin{abstract}
We present a customized 3D mesh Transformer model for the pose transfer task. As the 3D pose transfer essentially is a deformation procedure dependent on the given meshes, the intuition of this work is to perceive the geometric inconsistency between the given meshes with the powerful self-attention mechanism. Specifically, we propose a novel geometry-contrastive Transformer that has an efficient 3D structured perceiving ability to the global geometric inconsistencies across the given meshes. Moreover, locally, a simple yet efficient central geodesic contrastive loss is further proposed to improve the regional geometric-inconsistency learning. At last, we present a latent isometric regularization module together with a novel semi-synthesized dataset for the cross-dataset 3D pose transfer task towards unknown spaces. The massive experimental results prove the efficacy of our approach by showing state-of-the-art quantitative performances on SMPL-NPT, FAUST and our new proposed dataset SMG-3D datasets, as well as promising qualitative results on MG-cloth and SMAL datasets. It's demonstrated that our method can achieve robust 3D pose transfer and be generalized to challenging meshes from unknown spaces on cross-dataset tasks. The code and dataset are made available. Code is available: \url{https://github.com/mikecheninoulu/CGT}.
\end{abstract}

\vspace{-0.5cm}
\section{Introduction}

\begin{figure*}[!h] \small
    \centering
    \includegraphics[width=0.95\linewidth]{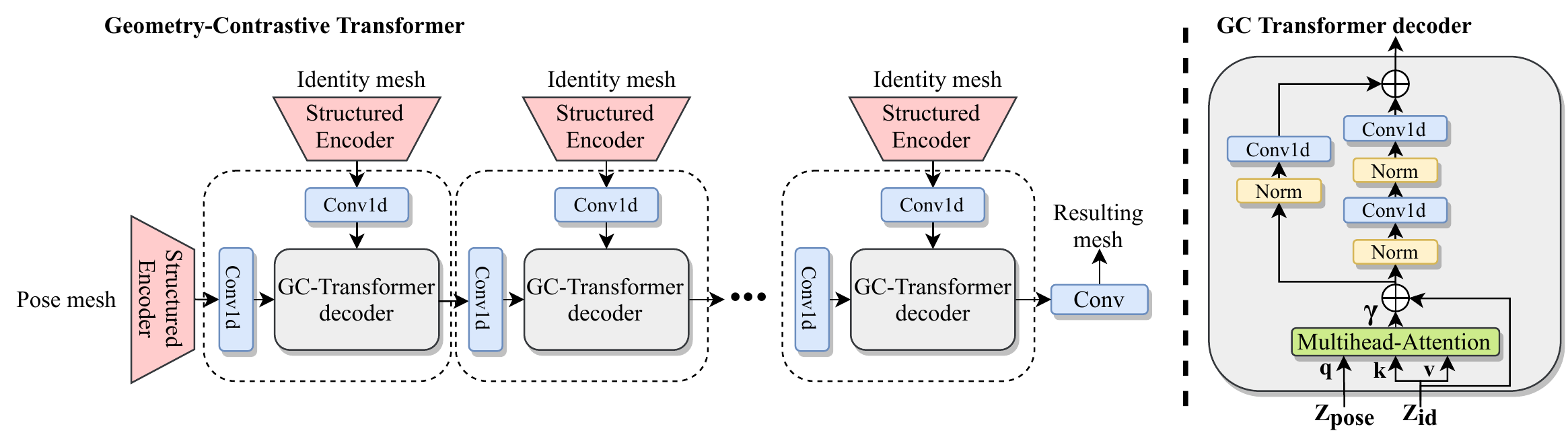}
    \caption{An overlook of our GC-Transformer. The left part is the whole architecture of the GC-Transformer. The right part illustrates the architecture details of one GC-Transformer decoder. The GC-Transformer borrows the idea from the work of \cite{transformer2d} but is extensively extended to 3D data processing tasks for both the encoders and decoders.}
    \vspace{-0.4cm}
    \label{fig:Network}
\end{figure*}
Pose transfer, applying the desired pose of a source mesh to a target mesh, is a promising and challenging task in 3D computer vision, which can be widely applied to various industrial fields. However, existing methods ~\cite{NPT,LIMP,Unsupervised,AniFormer} can only perform well within given datasets of synthesized/known pose and shape space, and fail to be generalized to other unknown spaces with robust performances, which severely limits the further real-world implementations. 

To achieve robust performances on unknown latent spaces and other domains as shown in Fig.~\ref{fig:1}, we propose a novel Transformer network targeting generalized 3D mesh pose transfer. Specifically, a novel geometry-contrastive Transformer with geometrically structured encoders is designed that aims to enhance the identity mesh representation under the guidance of the pose mesh with their \emph{global geometric contrasts}. Locally, we introduce a novel central geodesic contrastive loss to improve the geometric representation by considering the \emph{regional contrast of all the geodesic directions} of each vertex as back-propagation gradients. Furthermore, we present a latent isometric regularization module to stabilize the unreliable performance of cross-dataset pose transfer problems.

Moreover, we present a new 3D mesh dataset, i.e., SMG-3D, for quantitatively evaluating the 3D pose transfer with unknown spaces. The SMG-3D is based on daily spontaneously performed body gestures with more plausible and challenging body movements and different than those well-performed poses \cite{AMASS,DFAUST}. We use a semi-synthesis way to build the dataset to provide necessary GT meshes for training and validating. Our SMG-3D dataset can be jointly combined with other existing body mesh datasets for cross-dataset qualitative analysis.

A natural question to ask is: why not simply use purely synthesized meshes to train and evaluate the model? The short answer is that models trained on purely synthesized meshes will fail in the cross-dataset task. Indeed, using mesh synthesizing models like the SMPL series~\cite{SMPL, SMAL, SMPLX} can synthesize unlimited poses that can cover the whole latent space, or a large-scale dataset AMASS \cite{AMASS} to eliminate the inconsistencies with unknown dataset space. However, in practice, even for a small dataset FAUST with only 10 pose categories, it takes more than 26 hours to train a model~\cite{LIMP} to fully learn the latent space. Thus, due to the staggering variability of poses and movements, it's not feasible to train the model with synthesized samples covering the whole pose space. It's desirable that a model can be directly generalized to unknown latent spaces in a more efficient way. To this end, we propose the SMG-3D dataset to tackle the cross-dataset learning issue. It can provide challenging latent distribution allocates on natural and plausible body poses with occlusions and self-contacts instead of well-posed body moves like AMASS~\cite{AMASS}, which \textit{could advance the research to real-world scenarios one step further}. 

To summarize, our contributions are as follows:
\begin{itemize}[leftmargin=*]
\item A novel geometry-contrastive Transformer of positional embedding free architectures with state-of-the-art performances on the challenging 3D pose transfer task.
\item A simple and efficient central geodesic contrastive loss that can further improve the geometric learning via preserving the direction gradient of the 3D vertices.
\item A challenging 3D human body mesh dataset (i.e., SMG-3D) providing unknown space of naturally plausible body poses with challenging occlusions and self-contacts for the cross-dataset qualitative evaluation.
\item  A new latent isometric regularization module for adapting to challenging unknown spaces on cross-dataset tasks. 
\end{itemize}
\section{Related work}
\begin{figure*}[t!h] \small
    \centering
    \includegraphics[width=\linewidth]{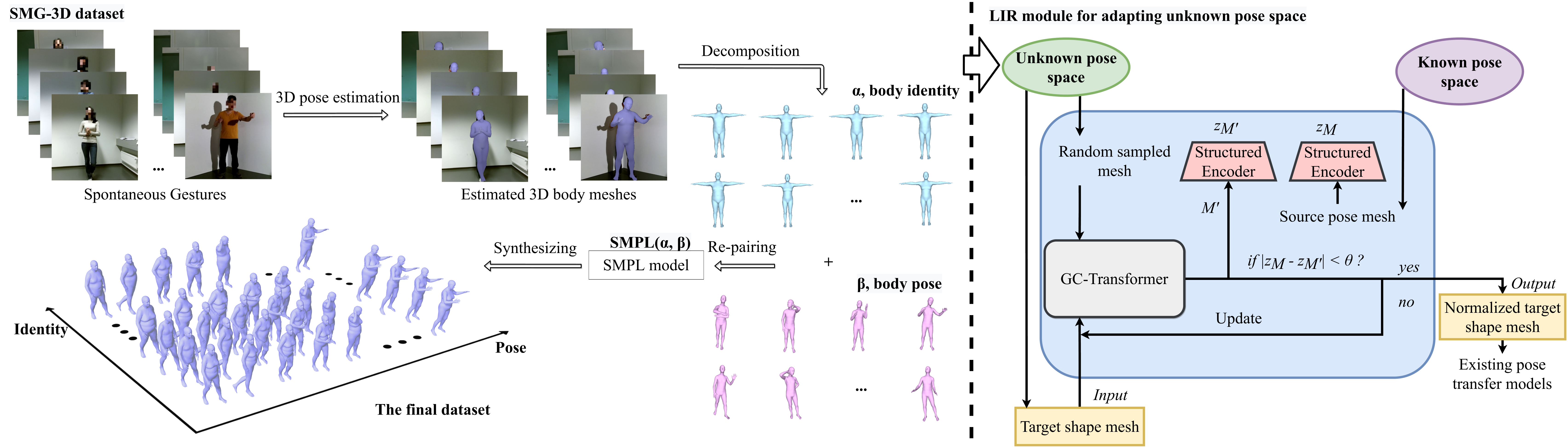}
    \caption{Left: an overlook of our semi-synthesized 3D mesh body gesture dataset SMG-3D. It is a 3D dataset with a pose space that fits the real-world dataset pose distribution, including naturally and spontaneously performed body movements in daily communication with challenging occlusions and self-contacts. Right: the architecture of proposed Latent Isometric Regularization (LIR) module for unknown latent space learning.}
    \vspace{-0.4cm}
    \label{fig:SMG-3D}
\end{figure*}

\noindent \textbf{3D Mesh Deformation Transfer.} Deformation transfer aims to generate a new 3D shape with a given pair of source poses and target shapes. Even though existing methods~\cite{3dcode,dt} could bring impressive deformation results, the superb performances largely rely on the given correspondences of the source and target meshes, which limits their generalization ability. Some disentanglement-based methods like~\cite{Unsupervised, LIMP, IEPGAN} tried to decompose meshes into shape and pose factors and achieve pose transfer as a natural consequence. However, extra constraints on the datasets are still needed.
\begin{table}[h]
\label{tab:comparison}
\caption{A comparison of our GC-Transformer with other 3D Transformer variants.}
\setlength{\belowcaptionskip}{-6.0cm}
\resizebox{1\linewidth}{!}{%
\begin{tabular}{@{}ccccc@{}}
\toprule
Model & \begin{tabular}[c]{@{}c@{}}Vertex \\ operator\end{tabular} & \begin{tabular}[c]{@{}c@{}}Vertex \\ topology\end{tabular} & Processed mesh size & Mesh type \\ \midrule
Vanilla TFM & MLP & Damaged & - & - \\ \midrule
\begin{tabular}[c]{@{}c@{}}METRO \\ \cite{metro} \end{tabular} & \begin{tabular}[c]{@{}c@{}}Positional \\ embedding\end{tabular} & \begin{tabular}[c]{@{}c@{}}Preserved, \\ high cost\end{tabular} & \begin{tabular}[c]{@{}c@{}}Down-sampled \\ from 6890 to 431\end{tabular} & \begin{tabular}[c]{@{}c@{}}Pseudo \\ (post-process)\end{tabular} \\ \midrule
\begin{tabular}[c]{@{}c@{}}PolyGen \\ \cite{polygen}\end{tabular}  & \begin{tabular}[c]{@{}c@{}}Pointer \\ embedding\end{tabular} & \begin{tabular}[c]{@{}c@{}}Preserved, \\ high cost\end{tabular} & \begin{tabular}[c]{@{}c@{}}Filter meshes larger\\ than 800 vertices\end{tabular} & Real mesh \\ \midrule
\textbf{GC-Transformer (Ours)} & \textbf{\begin{tabular}[c]{@{}c@{}}Depth-wise\\ 1D Conv\end{tabular}} & \textbf{\begin{tabular}[c]{@{}c@{}}Preserved, \\ no cost\end{tabular}} & \textbf{\begin{tabular}[c]{@{}c@{}}Original size \\ such as 6890\end{tabular}} & \textbf{Real mesh} \\ \bottomrule
\end{tabular}}
\vspace{-0.3cm} 
\label{tab:transformer}
\end{table}

\noindent\textbf{Deep Learning for Geometric Representation.} PointNet~\cite{pointnet} and PointNet++~\cite{pointnet++} have become common-use frameworks that can work directly on sparse and unorganized point clouds. After that, mesh variational autoencoders~\cite{vaelimp,vae2} were also proposed to learn mesh embedding for shape synthesis but they are under a strong condition that the shape of target objects should be given as prior. On the other hand, there is a trend to utilize to the self-attention mechanism of Transformers for structural geometric information learning. However, as shown in Table \ref{tab:transformer}, those preliminary works \cite{metro,polygen,pointTransformer} tried to directly encode the vertex topological structures with computationally demanding embeddings, thus can only handle small-size meshes. In this work, our GC-Transformer is completely different and implements depth-wise 1D Convolution instead of any computational embedding to preserve vertex topological structures thus freely handles LARGE meshes with fine-grained details at no cost, which could boost efficient implementations of Transformer frameworks in 3D fields.

\noindent\textbf{Cross-Dataset 3D Pose Transfer.} There is few 3D mesh dataset suitable for the pose transfer task. Though many techniques and body models have been developed for 3D data analysis such as SMPL series~\cite{SMPL, SMPLH, SMPLX, SMAL}, as well as various 3D human body datasets \cite{FAUST, DFAUST,MG-cloth, SMPLX, AMASS}, they are all originally designed for other tasks such as scan registration, recognition, or shape retrieval. Thus, the poses in those datasets are all exaggerated and perfectly posed actions, for instance, to ensure the quality of the scan registration. However, the latent space distribution of real ones with occlusion and self-contacts can differ widely. Besides, few of the existing datasets can be parameterized and manipulated in the latent space towards desired poses, thus no standard GT is available for the training and the quantitative evaluation. Existing methods~\cite{LIMP} could merely use approximations such as geodesic preservation as substitutes.

\section{Methodology}
We define a 3D parametric mesh as $M(\alpha, \beta)$, where $\alpha$, $\beta$ denote the parameters of identity (i.e., shape) and pose. Let $M^1(\alpha_{pose}, \beta_{pose})$ be the mesh with the desired pose for style transfer and $ M^2(\alpha_{id}, \beta_{id})$ be the mesh with its identity to preserve. Then the polygon mesh $M' (\alpha_{id}, \beta_{pose})$ is the target to generate. The goal of pose transfer is to learn a deformation function $f$ which takes a pair $M^1$ and $M^2$ and produces a new mesh ${M'}$, so that the geodesic preservation of the resulting mesh ${M'}$ is identical to the source one~$M^2$ and the pose style is identical to $M^1$.
\begin{equation}
\label{Definemath}
\begin{aligned}
f(M^1(\alpha_{id}, \beta_{id}),M^2(\alpha_{pose}, \beta_{pose})) = {M'}(\alpha_{id},\beta_{pose}).
\end{aligned}
\end{equation}

Below, we will first introduce how to use the Transformer architecture-based model, called Geometry-Contrastive Transformer (GC-Transformer) for learning the deformation function $f$, then the Central Geodesic Contrastive (CGC) loss for detailed geometric learning, and at last, the Latent Isometric Regularization (LIR) module for robust pose transfer on cross-dataset tasks.
\vspace{-0.05cm}

\subsection{Geometry-Contrastive Transformer}
\label{sec:GCtransformer}
An overview of the GC-Transformer is depicted in Fig.~\ref{fig:Network}. Our GC-Transformer consists of two key components, one is a structured 3D mesh feature encoder and the other one is a Transformer decoder.

\noindent \textbf{Structured 3D Encoder.} As mentioned, existing 3D Transformers needs computationally demanding embeddings to encode vertex positions, thus in practice can only process ‘toy’ meshes. Inspired by NeuralBody \cite{neuralbody} that uses structured latent codes to preserve the vertex topology, we modify the conventional PointNet~\cite{pointnet} into structured 3D encoders to capture the vertex topology by implementing depth-wise 1D convolution instead of redundant positional embeddings commonly used in conventional Transformers. Meanwhile, we replace the batch normalization layers into Instance Normalization~\cite{instance} layers to preserve the instance style which is widely used on style transfer tasks~\cite{styletransfer, spatiallynormalization}. The resulting latent embedding vector $Z$ with dimension $N_{latent}$ from the encoder will be dimensionally reduced with 1D convolution and fed into the following GC-Transformer decoder. In this way, LARGE meshes with fine-grained details can be handled freely at no cost by our GC-Transformer while preserving the vertex structures.

\begin{figure*}[!h]
    \centering
    \includegraphics[width=\linewidth]{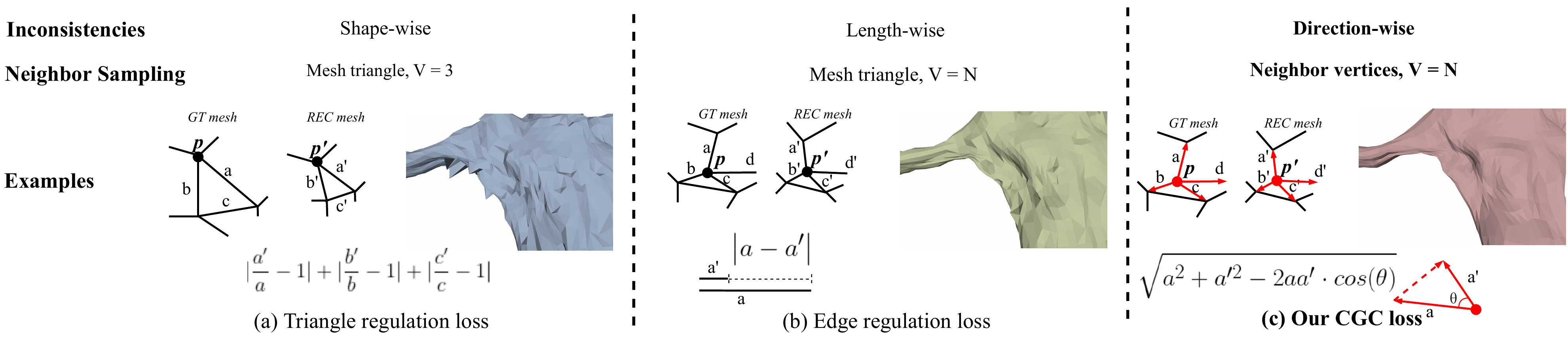}
    \caption{A comparison of different losses for both the neighbor vertex sampling strategy and the local inconsistency. Our CGC loss considers the inconsistencies of all the geodesic directions at each vertex, so that direction gradients can be preserved in the back-propagation. Results show that CGC loss can make the local details more tight and realistic.}
    \vspace{-0.4cm}
    \label{fig:cdcloss}
\end{figure*}

\noindent\textbf{GC-Transformer Decoder.} We encourage readers to refer to \cite{transformer2d} for a standard Transformer structure, which achieve state-of-the-art results on many tasks such as \cite{li2021mhformer,yang2021transformer}. We propose the GC-Transformer decoder that inherits the classical structure with customized designs for 3D meshes. The structure of the GC-Transformer decoder is shown in Fig.~\ref{fig:Network}.

The core difference between the GC-Transformer and a standard Transformer is the design of the multihead self-attention. To learn the correlations between the given meshes for geometric deformation, the model should be able to perceive the geometric information from the two meshes. Thus, we make the inputs of a GC-Transformer as the latent embedding vectors of \emph{two meshes} instead of a single input like the classical Transformer. Besides, as it's a style transfer task, we utilize the Instance Norm introduced by~\cite{styletransfer} as our normalization layers. At last, to preserve the structural information of 3D data, the MLP layers are replaced with 1D Convolutional layers. 

We denote the latent embedding vectors of the pose mesh and identity mesh from the encoders as $Z_{pose}$ and $Z_{id}$ respectively. We feed the two embedding vectors into different 1D convolution layers to generate the representations $\mathbf{qkv}$ for the standard multihead self-attention \cite{transformer}. The query $\mathbf{q}$ is from $Z_{pose}$, and the value $\mathbf{v}$ and key $\mathbf{k}$ are from $Z_{id}$. Then, the attention weights $A_{i,j}$ based on the geometric pairwise similarity between two elements of $\mathbf{q}$ and $\mathbf{k}$ is given with the following formula:
\begin{equation}
\label{geometric correlation}
\mathbf{A}_{i,j} = \frac{exp(\mathbf{q}_{i}\mathbf{k}_{j})}{\sum_{i=1}^{n}exp(\mathbf{q}_{i}\mathbf{k}_{j})}.
\end{equation}
After this, a matrix multiplication between $v$ and the transpose of $\mathbf{A}$ is conducted to perceive the geometric inconsistency between meshes. Finally, we weigh the result with a scale parameter $\gamma$ and conduct an element-wise sum operation with the original latent embedding $Z_{pose}$ to obtain the refined latent embedding $Z'_{pose}$,
\begin{equation}
\label{updateembedding}
Z'_{pose} =\gamma \sum_{i=1}^{n}\left ( \mathbf{A}_{i,j}\mathbf{v}_{i} \right ) + Z_{pose},
\end{equation}
where $\gamma$ is initialized as 0 and updated gradually during the training with gradients. The obtained $Z'_{pose}$ is followed by typical Transformer operators as introduced above Fig.~\ref{fig:Network} with a convolutional layer and Tanh activation, generating the final output ${M'}$. Please refer to the supplementary materials for more implementing details. 

In such a crossing way, the geometric-perceived feature code can consistently be rectified by the original identity mesh and its latent embedding representations. Note that, different than previous attention-based modules~\cite{nonlocal,xinggan,styletransfer,tang2020bipartite}, our GC-Transformer could not only compute the pair-wise correlations and contrasts in a crossing-mesh way but also could fully preserve the local geometric details with the residual layer. Most importantly, our GC-Transformer is designed for 3D mesh processing which has never been attempted in these works. Note that input mesh vertices are all shuffled randomly to ensure the network is vertex-order invariant.

\subsection{Central Geodesic Contrastive Loss}

Most of the existing 3D mesh representation learning losses, such as triangle regularization loss, edge loss, Chamfer loss and Laplacian loss \cite{pixel2mesh, NPT, 3dcode, Laploss, fully} all repeal the gradient of the direction information of 3D vertices. They only compare the scalar (or weak vector) differences of the mesh vertices such as one-ring geodesic lengths to construct the objective function, while the convexity of the mesh surface  containing rich directional gradient information is not utilized. To this end, inspired by the superb performances of central difference convolution \cite{cdc,fas,mmcdc} that considers the directional difference of depth space, we suggest to compare the vector differences of the vertex topology by proposing a simple yet efficient central geodesic contrastive loss as below:
\begin{equation}
\label{cgcloss}
\begin{aligned}
\mathcal{L}_{contra} = \frac{1}{V}\sum_{\mathbf{p}}\sum_{\mathbf{u}\in \Gamma (\mathbf{p})}\sqrt{u_{M'}^{2} + u_{M}^{2} -2u_{M'}u_{M}\cdot cos(\theta ) },
\end{aligned}
\end{equation}
where $\Gamma (\mathbf{p})$ denotes the neighbor edges of vertex $\mathbf{p}$ and $V$ is the total vertex number of the mesh. $u_{M}$ denotes the edge of mesh $M$ and $\theta$ denote the included angle of the edges of $u_{M}$ and $u_{M'}$. In practice, $\mathcal{L}_{contra}$ can be easily calculated by taking the vector difference of $u_{M}$ and $u_{M'}$ within the coordinate of each vertex $p$ and divided by the total vertex number as a global normalization.

Our CGC loss has three improvements compared to existing losses: 1) the full inconsistencies of vertex vectors are calculated to preserve the direction gradient; 2) each direction of the vertex is separately considered instead of a simple sum-up; 3) the sampling methods of the neighbor vertices of $\mathbf{p}$ in Eq.~\eqref{cgcloss} is different: the CGC loss samples all the vertices connected to $\textbf{p}$ resulting in a flexible $N$ neighbor vertices while \cite{pixel2mesh, 3dcode} are within the mesh triangle of vertex $\mathbf{p}$ and fixed to 3. Please refer to Fig.~\ref{fig:cdcloss} for a better understanding. A point-wise $L2$ reconstruction loss of mesh vertices can only capture the absolute distance in the coordinate space. Contrastively, our CGC loss captures the inconsistencies of all the geodesic directions at each vertex, so that direction gradients can be preserved in the back-propagation. Note that our CGC loss is similar to Laplacian loss but can preserve full vector differences without Laplacian normalization, thus is not only limited to smooth surfaces. As shown in Fig.~\ref{fig:cdcloss}, our CGC loss could offer additional strong supervision especially in tightening the output mesh surface.

\noindent\textbf{Overall Objective Function.} With our proposed CGC loss, we define the full objective function as below:
\begin{equation}
\label{loss}
\mathcal{L}_{full} =\lambda_{rec}\mathcal{L}_{rec} + \lambda_{edge}\mathcal{L}_{edge} + \lambda_{contra}\mathcal{L}_{contra},
\end{equation}
where $\mathcal{L}_{rec}$, $\mathcal{L}_{edge}$ and $\mathcal{L}_{contra}$ are the three losses used as our full optimization objective, including reconstruction loss, edge loss and our newly proposed CGC loss. $\lambda$ is the corresponding weight of each loss. In Eq.~\eqref{loss}, reconstruction loss $\mathcal{L}_{rec}$ is the point-wise L2 distance and the edge loss~\cite{3dcode} is an edge-wise regularization between the GT meshes and predicted meshes.

\subsection{Cross-Dataset Pose Transfer} 
Although existing pose transfer methods can deal with fully synthesized/known pose space, they fail to have a robust performance on the pose space that is different from the training one. To facilitate the 3D analysis of human behaviors to real-world implementations, we propose a new SMG-3D dataset as well as a LIR module towards the cross-dataset issue.

\noindent\textbf{A New SMG-3D Dataset.} The main contribution of the SMG-3D dataset is providing an alternative benchmark towards cross-dataset tasks by providing standard GTs under a challenging latent pose distribution (unlike perfectly synthesized/performed known distributions). As shown in Fig.~\ref{fig:SMG-3D}, SMG-3D is derived from an existing 2D body pose dataset called SMG dataset~\cite{SMG} that consists of spontaneously performed body movements with challenging occlusions and self-contacts. Specifically, we first adopt the 3D mesh estimation model STRAPS~\cite{STRAPS} to generate the 3D mesh estimations from the original 2D images of SMG. Then, we select 200 poses and 40 identities as templates to form the potential pose space and optimize them by Vposer~\cite{SMPLX}. At last, the generated 3D meshes are decomposed into numerical registrations as latent parameters which are paired to synthesize the resulting 8,000 body meshes via the SMPL model~\cite{SMPL}, each with 6,890 vertices. Compared to synthesized/well-performed meshes, our in-the-wild 3D body meshes are more practical and challenging with the large diversity and tricky occlusions for providing the unknown latent space. Please check more about our dataset in the supplementary materials.

\begin{figure*}[!h] \small
\centering
    \includegraphics[width=\linewidth]{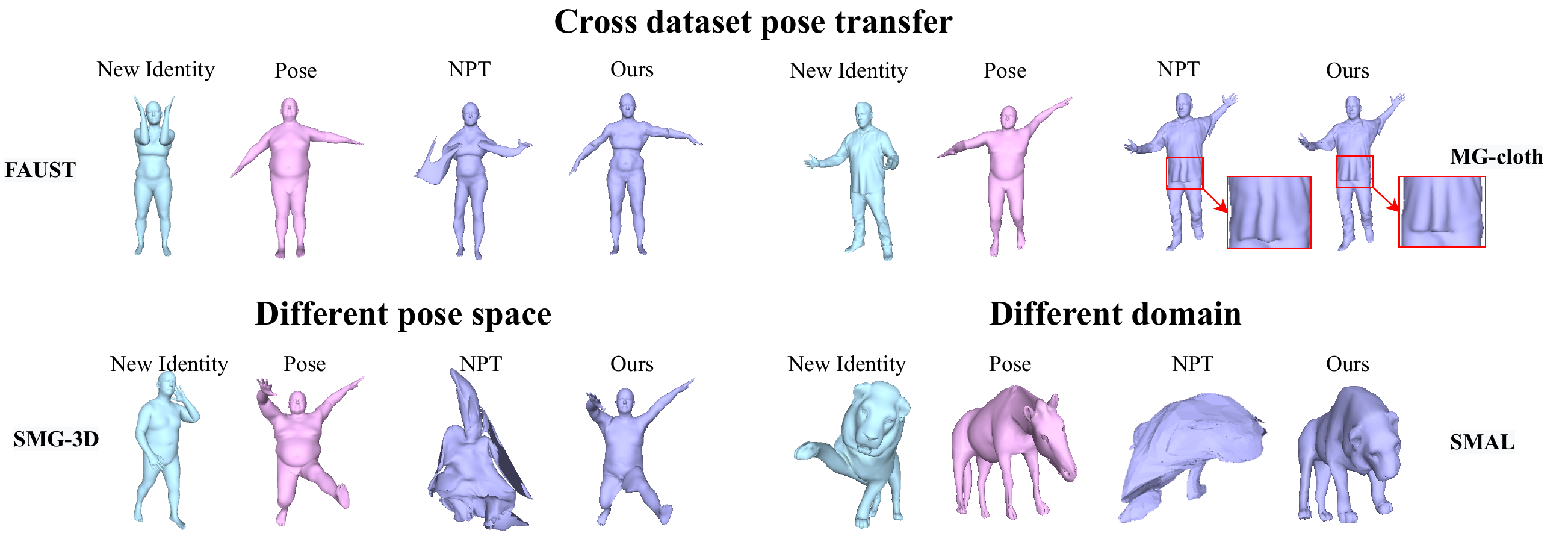}
    \caption{Cross-dataset qualitative results compared with NPT~\cite{NPT} on four different datasets. FAUST, SMG-3D and MG-cloth could conduct the pose transfer directly with a model trained on SMPL-NPT.}
        \vspace{-0.4cm}
    \label{fig:qualitiveunseen}
\end{figure*}

\noindent\textbf{Latent Isometric Regularization Module.} When the poses and shapes are from unknown latent spaces, existing methods suffer from degeneracy in varying degrees (see Fig.~\ref{fig:qualitiveunseen}). We address this issue by introducing the LIR module as shown in Fig.~\ref{fig:SMG-3D} right part, that can aggregate the data distribution of target set and source set. The LIR can be \emph{stacked to existing standard models} to enhance the cross-dataset performance. Specifically, the difference between the two datasets is obtained by comparing the latent pose codes $z_M$ and $z_{M'}$ of the shape mesh $M'$ from the target set and the pose mesh $M$ from the source dataset. The target shape mesh will be fed into GC-Transformer along with another randomly sampled mesh from the target set to obtain a newly generated mesh $M'$. This will be iteratively executed until the latent pose code difference $z_{M'}$ and $z_{M}$ converges to less than $\theta$, resulting in a normalized target set. In this way, the latent pose distribution of the target set will be regulated while its isometric information can still be preserved. Essentially, our LIR module serves as a domain adaptive normalization to warm-up the unknown target set to better fit the model trained on the source pose space. 


\begin{figure*}[!h] \small
    \centering
    \includegraphics[width=0.8\linewidth]{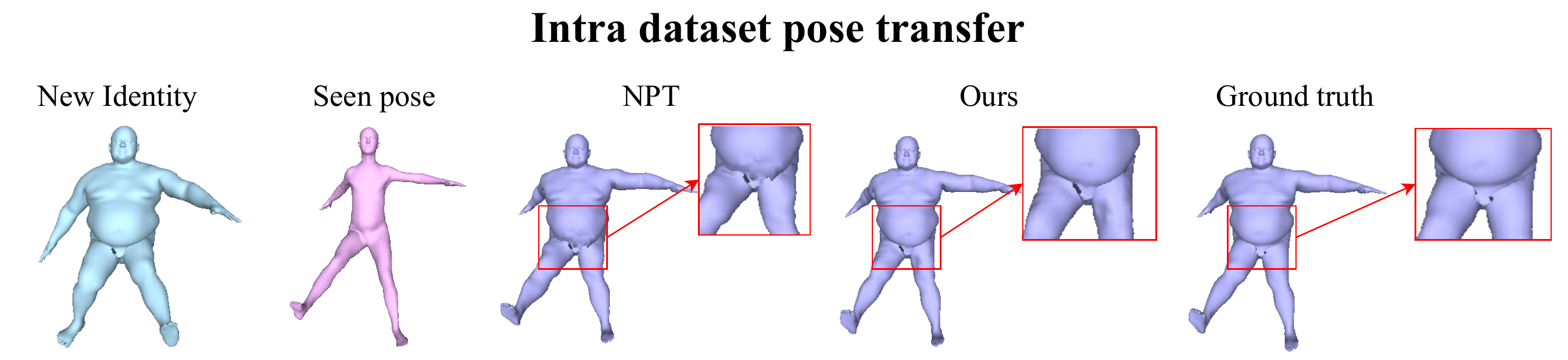}
    \caption{Intra-dataset qualitative results compared with NPT \cite{NPT} on the SMPL-NPT dataset. With satisfying visual effects of both compared methods, our GC-Transformer have a better representation ability in geometry details.}
        \vspace{-0.4cm}
    \label{fig:qualitiveseen}
\end{figure*}

\section{Experiments}

\begin{table}[] \small
\caption{Intra-dataset performances on SMG-3D and SMPL-NPT datasets. ``NPT MP'' stands for NPT model with max pooling layers. Note that the ``unseen'' setting is still within the same dataset with similar data distributions.}
\centering
	\resizebox{1\linewidth}{!}{%
\begin{tabular}{@{}ccccccc@{}}
\toprule
\multirow{3}{*}{\begin{tabular}[c]{@{}c@{}}PMD$\downarrow$\\ $ (\times 10^{-4})$\end{tabular}} & \multicolumn{3}{c}{Seen} & \multicolumn{3}{c}{Unseen} \\ \cmidrule(l){2-7} 
 & \begin{tabular}[c]{@{}c@{}}NPT-MP\\ \cite{NPT}\end{tabular} & \begin{tabular}[c]{@{}c@{}}NPT\\ \cite{NPT}\end{tabular}& \begin{tabular}[c]{@{}c@{}}GC-\\ Transformer~\end{tabular} & \begin{tabular}[c]{@{}c@{}}NPT-MP\\ \cite{NPT}\end{tabular}& \begin{tabular}[c]{@{}c@{}}NPT\\ \cite{NPT}\end{tabular} & \begin{tabular}[c]{@{}c@{}}GC-\\ Transformer\end{tabular} \\ \midrule
 SMG-3D & 70.3 & 62.1 & \textbf{30.7} & 120.3 & 94.6 &\textbf{52.8}   \\
 SMPL-NPT & 2.1 & 1.1 & \textbf{0.6} & 12.7 & 9.3 & \textbf{4.0}\\ \bottomrule
\end{tabular}}
\label{Tab:intradataset}
\vspace{-0.4cm}
\end{table}

\subsection{Datasets}
\noindent \textbf{SMPL-NPT} \cite{NPT} dataset contains 24,000 synthesized body meshes with the SMPL model~\cite{SMPL} by sampling in the parameter space. For training, 16 different identities and 400 different poses are randomly selected and made into pairs as GTs. For testing, 14 new identities are paired with those 400 poses and 200 new poses as ``seen'' and ``unseen'' sets. Note that the ``unseen'' poses are sampled within \textbf{the same parameter distribution} as the ``seen'' poses, thus still in the \textit{same/known latent pose space}.

\noindent \textbf{SMG-3D} \cite{SMG} dataset contains 8,000 pairs of naturally plausible body meshes of 40 identities and 200 poses, 35 identities and 180 poses are used as the training set. The rest 5 identities with the 180 poses and the other 20 poses are used for ``seen'' and ``unseen'' testing. Note that both SMPL-NPT and SMG-3D provide GT meshes so that they can be used for cross-dataset quantitative evaluation.

\noindent \textbf{FAUST}~\cite{FAUST} dataset consists of 10 different human subjects, each captured in 10 poses. The FAUST mesh structure is similar to SMPL with 6,890 vertices.

\noindent \textbf{MG-Cloth}~\cite{MG-cloth} dataset contain 96 dressed identity meshes with different poses and clothes. The MG-cloth meshes contain way more vertices (above 27,000), which is more challenging for more fine-grained geometry details. Note that meshes in FAUST and MG-cloth are not parameterized SMPL models so geodesic-based approximations~\cite{GIH} is always used for evaluation in previous works. 

\noindent \textbf{SMAL}~\cite{SMAL} animal dataset is based on a parametric articulated quadrupedal animal model and we adopted it to synthesize the training and testing datasets.

\subsection{Intra-Dataset Pose Transfer Evaluation}
Firstly, we evaluate the intra-dataset pose transfer performance of our GC-Transformer on the SMPL-NPT and SMG-3D. Given the GT meshes, we follow~\cite{NPT} to adopt Point-wise Mesh Euclidean Distance (PMD) as the evaluation metric:
\begin{equation}
\label{PMD}
PMD = \frac{1}{|V|} \sum_{\mathbf{v}}\left \| M_{\mathbf{v}}-M'_{\mathbf{v}} \right \|_{2}^{2}.
\end{equation}
where $M_{\mathbf{v}}$ and $M'_{\mathbf{v}}$ are the point pairs from the GT mesh $M$ and generated one $M'$. The final experimental results can be found in Table~\ref{Tab:intradataset}. For both settings of the SMPL-NPT: ``seen'' and ``unseen pose'', our GC-Transformer significantly outperforms compared SOTA methods by more than 45\% and 55\% with PMD ($\times 10^{-4}$) of: 0.6 and 4.0 vs. 1.1 and 9.3. We denote PMD ($\times 10^{-4}$) as PMD for simplicity in the following. On our SMG-3D dataset, our network again yields the best performance among other methods with PMD of (30.7 and 52.8). As shown, the SMG-3D is more challenging than the SMPL-NPT dataset with way higher PMD values for all the models. Compared to the fully synthesized dataset SMPL-NPT, the poses in SMG-3D are more realistic as they contain many occlusions and self-contacts. The distribution of the poses in the latent space is significantly uneven and discontinuous while the poses synthesized in the SMPL-NPT dataset are way easier with less noise.
\begin{figure*}[!h] \small
    \centering
    \includegraphics[width=0.9\linewidth]{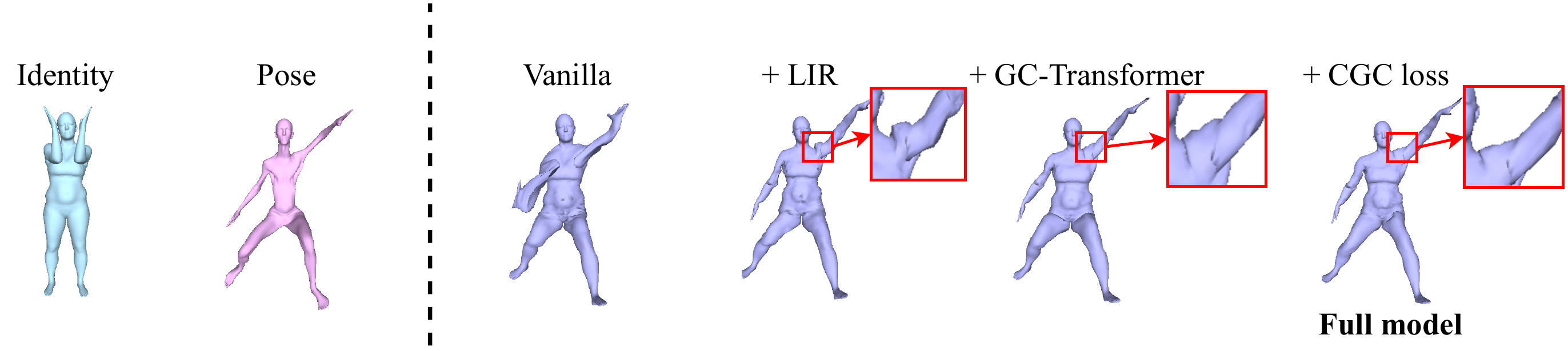}
    \caption{Ablation study by progressively enabling each component. The rightmost mesh is from the full GC-Transformer.}
    \vspace{-0.2cm}
    \label{fig:ablationstudy}
\end{figure*}

\subsection{Generalized Pose Transfer Evaluation}
\begin{table}[] \small
\caption{Cross-dataset performances on FAUST dataset. Because we use the raw meshes of FAUST and there is no GT, geometric approximations are used for evaluation.}
\centering
	\resizebox{1\linewidth}{!}{%
\begin{tabular}{@{}cccc@{}}
\toprule
\multicolumn{4}{c}{Disentnaglement Error}                 \\ \midrule
\begin{tabular}[c]{@{}c@{}}VAE\\ ~\cite{vaelimp}\end{tabular} & \begin{tabular}[c]{@{}c@{}}LIMP-Euc\\ \cite{LIMP}\end{tabular} & \begin{tabular}[c]{@{}c@{}}LIMP-Geo\\ \cite{LIMP}\end{tabular} & \begin{tabular}[c]{@{}c@{}}GC-\\ Transformer\end{tabular} \\ \midrule
7.16 & 4.04 & 3.48 & \textbf{0.11} \\ \bottomrule
\end{tabular}}
\label{Tab:geocorss}
\vspace{-0.4cm}
\end{table}

\noindent \textbf{Cross-Dataset Pose Transfer with Same Pose Space}. We extent the setting to cross-datasets by training the model on SMPL-NPT dataset and directly conduct the pose transfer on the unseen meshes from FAUST and MG-cloth datasets. As shown in Fig.~\ref{fig:qualitiveunseen} first line, NPT might fail when the target pose is not within the training latent space while our method can still perform well. Since there is no GT available here, we adopt the disentanglement error of the pose transfer task illustrated in \cite{LIMP} as the metrics, see \cite{LIMP} for more details. In Table \ref{Tab:geocorss}, we report the performances of GC-Transformer and state-of-the-art models on FAUST. Compared to \cite{LIMP} trained with the preservation of geodesic distances, ours significantly outperforms \cite{LIMP}: 0.23 vs. 3.48. As expected, the preservation of geodesic distances \cite{LIMP} can only serve as the approximation of GTs.

\begin{table}[] \small
\caption{Cross-Dataset performances with standard GTs as metrics. Our LIR module can be stacked to existing models and robustly improve the performances on unknown spaces.}
\centering
	\resizebox{1\linewidth}{!}{%
\begin{tabular}{@{}ccccc@{}}
\toprule
\multicolumn{2}{c}{Cross-dataset} & \multicolumn{3}{c}{PMD$\downarrow (\times 10^{-4})$} \\ \midrule
Training set & Testing set & \begin{tabular}[c]{@{}c@{}}NPT-MP\\ \cite{NPT}\end{tabular}  & \begin{tabular}[c]{@{}c@{}}NPT\\ \cite{NPT}\end{tabular}  & \begin{tabular}[c]{@{}c@{}}GC-\\ Transformer\end{tabular} \\ \midrule
\multirow{3}{*}{SMPL-NPT} & \textit{SMPL-NPT} & \textit{12.7} & \textit{9.3} & \textit{4.0} \\ \cmidrule(l){2-5} 
 & SMG-3D wo/LIR & 321.4 & 240.1 & 178.7 \\
 & SMG-3D w/LIR & 132.3 & 121.4 & \textbf{79.2} \\ \midrule
\textit{SMG-3D} & \textit{SMG-3D} & \textit{120.3} & \textit{94.6} & \textit{52.8} \\ \bottomrule
\end{tabular}}
\label{Tab:GTcross}
 \vspace{-0.2cm}
\end{table}

\noindent \textbf{Cross-Dataset Pose Transfer with Different Pose Space}. In this part, we quantitatively analyze the cross-dataset performance between different latent spaces of SMPL-NPT and SMG-3D datasets by using GTs as metrics. As shown in Table~\ref{Tab:GTcross}. We directly use the model trained on SMPL-NPT to conduct the pose transfer on the meshes from SMG-3D. The performance of the GC-Transformer (PMD 79.2 and 178.7) keeps outperforming compared methods (PMD 121.4 and 240.1) as presented in Table~\ref{Tab:GTcross}. It can be seen that, by adopting our LIR module, all the models can effectively improve the performances which proves its efficiency, which also proves that the inconsistency of the latent pose space affects the generalization of the pose transfer. 

\noindent \textbf{Efficacy of SMG-3D Dataset}. From Table~\ref{Tab:GTcross}, we observe that models trained on the synthesized SMPL-NPT dataset can perform well within the same pose space (first row of the table). However, when directly transferring the model to a unknown space like SMG-3D, the PMD dramatically drops down. This proves that a model trained with purely synthesized datasets cannot fit the space distribution of challenging real-world poses. In contrast, by introducing SMG-3D dataset, we can train the model with semi-synthesized data to better fit the pose space of the real-world one, as shown in the last line (PMD improved from 321.4 to 120.3 for NPT and 178.7 to 52.8 for our GC-Transformer). As indicated, a model that works on whole latent pose space is challenging which proves the necessity of our SMG-3D dataset.

\noindent \textbf{Pose Transfer on Different Domain}. In the end, we show the robust performance of GC-Transformer on animal pose transfer in Fig.~\ref{fig:qualitiveunseen}. Our model can be directly trained on SMAL dataset without further modification to adapt the non-human meshes, showing a strong generalizing ability.

\subsection{Ablation Study}
\begin{table}[t] \small
\caption{Effect of GC-Transformer. We evaluate the GP-transformer by varying its multihead-attention block number with the rest of the model untouched.}
\centering
\resizebox{0.8\linewidth}{!}{%
\begin{tabular}{@{}ccccc@{}}
\toprule
\multirow{2}{*}{Pose Source} & \multicolumn{4}{c}{PMD$\downarrow (\times 10^{-4})$}                                                          \\ \cmidrule(l){2-5}
                             & 1 block & 2 blocks & 3 blocks & 4 blocks \\ \midrule
Seen-pose                    & 1.4               & 1.0                & 0.9                &  \textbf{0.8}                \\
Unseen-pose                  & 7.3               & 4.9                & 4.9                &  \textbf{4.2}                \\ \bottomrule
\end{tabular}
}
\vspace{-0.2cm}
\label{Tab:nonlocal}
\end{table}

\begin{table}[t] \small
\caption{Effect of CGC loss. We validate the contribution of CGC loss by varying the weight of the CGC loss. As we can see, the CGC loss evidently improves the geometry learning by more than 20\%.}
\centering
\resizebox{0.8\linewidth}{!}{%
\begin{tabular}{@{}cccccc@{}}
\toprule
\multirow{2}{*}{Pose Source} & \multicolumn{5}{c}{PMD$\downarrow (\times 10^{-4})$}                \\ \cmidrule(l){2-6}
                             &$\lambda_{constra}{=} 0$  & 0.0005 & 0.001 & 0.005 & 0.05 \\ \midrule
Seen-pose                    & 0.83   &  \textbf{0.64}   & 0.84  & 0.92  & 1.13  \\
Unseen-pose                  & 4.21   &  \textbf{3.98}   & 4.27  & 4.55  & 4.71  \\ \bottomrule
\end{tabular}}
\vspace{-0.2cm}
\label{Tab:contrastiveloss}
\end{table}

Experiments are conducted to present the effectiveness of each proposed component on the SMPL-NPT dataset.

\noindent \textbf{Effect of GC-Transformer}. We vary the number of the multi-head attention blocks to show the effect brought by GC-Transformer in Table~\ref{Tab:nonlocal}. We observe that the proposed GC-Transformer with four multi-head attention blocks works the best. However, increasing the number of blocks further requires large computational consumption and reaches the GPU memory limits. Thus, we adopt four blocks as default in our experiments.

\noindent \textbf{Effect of CGC Loss}. We also validate the effect of CGC loss with different $\lambda_{constra}$ settings, as shown in Table~\ref{Tab:contrastiveloss}. It shows that it gains the best performance when $\lambda_{constra}$ is set as 0.0005, which proves that our CGC loss could effectively improve the geometric reconstruction results.

Lastly, we visually present the contributions made from each component in the GC-Transformer in Fig.~\ref{fig:ablationstudy}. We disable all the key components as a Vanilla model and enable each step by step. Compared to the Vanilla model, the GC-Transformer, LIR module and CGC loss can consistently improve geometric representation learning. All components can be easily stacked to other existing models.
\section{Conclusion}
We introduce the novel GC-Transformer, as well as the CGC loss that can freely conduct robust pose transfer on LARGE meshes at no cost which could be a boost to Transformers in 3D fields. Besides, the SMG-3D dataset together with LIR module can tackle the problem of unstable transferring performance as the cross-dataset benchmark. New SOTA results proves our framework's efficiency in robust and generalized pose transfer. The proposed components can be easily extended to other 3D data processing models.

\section{Acknowledgements}
This work was supported by the Academy of Finland for Academy Professor project EmotionAI (grant 336116, 345122) and project MiGA (grant 316765),  EU H2020 AI4Media (No. 951911) and Infotech Oulu projects, as well as the CSC-IT Center for Science, Finland, for computational resources.

\bibliography{aaai22}
\vfill

This supplementary material includes technical details and additional results that were not included in the main submission due to the lack of space. All results were obtained with exactly the same methodology as the one described in the main manuscript. We first provide more implementation details of our method. Next, we provide more details and examples of our SMG-3D dataset. Finally, we show more pose transfer results compared with the leading method. 

This document is also accompanied by a video, showing the full views of generated meshes. We strongly encourage readers to watch the accompanying video for better visualization.

\section{Implementation Details}
\label{sec:1}
\subsection{Network Architectures}

Our GC-Tranformer framework consists of three main parts: a LIR module, a mesh latent embedding encoder, and a Transformer decoder for pose transfer. We first introduce network structures of each component, then give the architectural parameters of the full model.

\textbf{LIR Module}. The full architecture of the LIR module is presented in Table \ref{Tab:LIR model}. Starting from the observation that the latent pose distribution in the target dataset can be uneven and can fluctuate greatly, we introduce the LIR module as a pre-processing strategy to normalize the distribution of the latent pose space. The module is trained separately from the geometry-contrastive Transformer and it is applied only on the given testing set. It can be regarded as a dataset-wise normalization procedure to the testing set.

\textbf{Structured Encoders}. The architecture of structured encoders is presented in Table \ref{Tab:encoder model}. The encoders are used to encode a given mesh into a latent embedding for further mesh generation with the following decoders. Note that in order to fit our model on non-SMPL mesh models (the vertex number of which is not equal to 6,890, such as MG-cloth of 27,000 vertices), we stack a max pooling layer to the end of the encoder. It can flexibly process meshes with different sizes into a certain one. Thus, a max pooling version is trained specifically on the SMPL-NPT dataset (with 6,890 inputs), then is evaluated on the MG-cloth dataset (with 27,554 as inputs). For the SMAL dataset, we train a model based on it and directly fix the vertex number to 3,889 both both training and testing sets.

\begin{table}[tbp]
\centering
\caption{Detailed architectural parameters for the LIR module. ``N'' stands for batch size and ``V'' stands for vertex number. The first parameter of conv1d is the  kernel size, the second is the stride size.}
\resizebox{\columnwidth}{!}{%
\begin{tabular}{cccc}
\toprule
Index & Inputs & Operation & Output Shape \\ \midrule
(1) & - & Identity mesh & N×3×V \\
(2) & (1) & Encoder & N×1024×V \\
(3) & (2)(2)(1) & GC-Transformer decoder 1 & N×1024×V \\
(4) & (3) & conv1d (1 × 1, 1) & N×512×V \\
(5) & (4)(4)(1) & GC-Transformer decoder 2 & N×512×V \\
(6) & (5) & conv1d (1 × 1, 1) & N×256×V \\
(7) & (6)(6)(1) & GC-Transformer decoder 3 & N×256×V \\
(8) & (7) & conv1d (1 × 1, 1) & N×3×V \\
(9) & (8) & Tanh & N×3×V \\ \bottomrule
\end{tabular}
}

\label{Tab:LIR model}
\end{table}

\begin{table}[tbp]
\caption{Detailed architectural parameters for the encoder. ``N'' stands for batch size and ``V'' stands for vertex number. The first parameter of conv1d is the kernel size, the second is the stride size.}
\centering
\resizebox{\columnwidth}{!}{%
\begin{tabular}{@{}cccc@{}}
\toprule
Index & Inputs & Operation & Output Shape \\ \midrule
(1) & - & Input mesh & N×3×V \\
(2) & (1) & conv1d (1 × 1, 1) & N×64×V \\
(3) & (2) & Instance Norm, Relu & N×64×V \\
(4) & (3) & conv1d (1 × 1, 1) & N×128×V \\
(5) & (4) & Instance Norm, Relu & N×128×V \\
(6) & (5) & conv1d (1 × 1, 1) & N×1024×V \\
(7) & (6) & Instance Norm, Relu & N×1024×V \\
(8) & (7) & Max pooling (for non-SMPL) & N×1024×V' \\ \bottomrule
\end{tabular}
}

\label{Tab:encoder model}
\end{table}

\textbf{GC-Transformer Decoder}. The network architecture of a GC-Transformer decoder is presented in Table~\ref{Tab:CGP block}. The normalization module structure is from \cite{NPT} presented in Table \ref{Tab:SPAdaIN block}, separately. The GC-Transformer decoder is used to conduct a cross-geometry attention of the given two meshes via their latent embedding.

\begin{table}[tbp]
\caption{Detailed architectural parameters for GC-Transformer decoder. ``N'', ``C'', and ``V'' stand for batch size, channel number, and vertex number, respectively. The first parameter of conv1d is he kernel size, the second is the stride size.}
\centering
\resizebox{\columnwidth}{!}{%
\begin{tabular}{cccc}
\toprule
Index & Inputs & Operation & Output Shape \\ \midrule
(1) & - & Identity Embedding & N×C×V \\
(2) & - & Pose Embedding & N×C×V \\
(3) & (1) & conv1d (1 × 1, 1) & N×C×V \\
(4) & (2) & conv1d (1 × 1, 1) & N×C×V \\
(5) & (3) & Reshape & N×V×C \\
(6) & (5)(4) & Batch Matrix Product & N×V×V \\
(7) & (6) & Softmax & N×V×V \\
(8) & (7) & Reshape & N×V×V \\
(9) & (2) & conv1d (1 × 1, 1) & N×C×V \\
(10) & (2)(8) & Batch Matrix Product & N×C×V \\
(11) & (10) & Parameter gamma & N×C×V \\
(12) & (11)(2) & Add & N×C×V \\
(13) & - & Pose Mesh & N×3×V \\
(14) & (12)(13) & Norm block & N×C×V \\
(15) & (14) & conv1d(1 × 1, 1), Relu & N×C×V \\
(16) & (14)(15) & Norm block & N×C×V \\
(17) & (16) & conv1d(1 × 1, 1), Relu & N×C×V \\
(18) & (12)(13) & Norm block & N×C×V \\
(19) & (18) & conv1d(1 × 1, 1), Relu & N×C×V \\
(20) & (17)(19) & Add & N×C×V \\ \bottomrule
\end{tabular}
}

\label{Tab:CGP block}
\end{table}

\begin{table}[tbp]
\caption{Detailed architectural parameters for Norm block. ``N'', ``C'', and ``V'' stand for batch size, channel number, and vertex number, respectively. The first parameter of conv1d is the kernel size, the second is the stride size.}
\centering
\begin{tabular}{cccc}
\toprule
Index & Inputs & Operation & Output Shape \\ \midrule
(1) & - & Pose Embedding & N×C×V \\
(2) & (1) & Instance Norm & N×C×V \\
(3) & - & Identity Mesh & N×3×V \\
(4) & (3) & conv1d (1 × 1, 1) & N×C×V \\
(5) & (3) & conv1d (1 × 1, 1) & N×C×V \\
(6) & (4)(2) & Multiply & N×C×V \\
(7) & (6)(5) & Add & N×C×V \\ \bottomrule
\end{tabular}

\label{Tab:SPAdaIN block}
\end{table}

Finally, we present the full model architecture in Table~\ref{Tab:fullmodel}. This is a GC-Transformer with four encoders and four GC-Transformer decoders. The settings of compared networks with three, two and one multihead self-attentions in the ablation study section can be obtained by removing modules from (3) to (12) in the GC-Transformer decoder 3, 2 and 4, respectively.

\begin{table}[tbp]
\caption{Detailed architectural parameters for the full model. ``N'' stands for batch size and ``V'' stands for vertex number. The first parameter of conv1d is the kernel size, the second is the stride size.}
\label{Tab:fullmodel}
\centering
\resizebox{\columnwidth}{!}{%
\begin{tabular}{@{}cccc@{}}
\toprule
Index & Inputs & Operation & Output Shape \\ \midrule
(1) & - & Identity Mesh & N×3×V \\
(2) & - & Pose Mesh & N×3×V \\
(3) & (1) & Encoder & N×1024×V \\
(4) & (2) & Encoder & N×1024×V \\
(5) & (3) & conv1d (1 × 1, 1) & N×1024×V \\
(6) & (4) & conv1d (1 × 1, 1) & N×1024×V \\
(7) & (6)(5)(1) & GC-Transformer decoder 1 & N×1024×V \\
(8) & (5) & conv1d (1 × 1, 1) & N×512×V \\
(9) & (7) & conv1d (1 × 1, 1) & N×512×V \\
(10) & (9)(8)(1) & GC-Transformer decoder 2 & N×512×V \\
(11) & (8) & conv1d (1 × 1, 1) & N×512×V \\
(12) & (10) & conv1d (1 × 1, 1) & N×512×V \\
(13) & (12)(11)(1) & GC-Transformer decoder 3 & N×512×V \\
(14) & (11) & conv1d (1 × 1, 1) & N×256×V \\
(15) & (13) & conv1d (1 × 1, 1) & N×256×V \\
(16) & (15)(14)(1) & GC-Transformer decoder 4 & N×256×V \\
(17) & (16) & conv1d (1 × 1, 1) & N×3×V \\
(18) & (17) & Tanh & N×3×V \\ \bottomrule
\end{tabular}
}

\end{table}

\subsection{Training Settings}

\begin{table}[tbp]
\caption{Sample settings for the SMPL-NPT and SMG-3D datasets. ``Used pairs'' stands for the sample sizes of training or testing sets. }
\label{Tab:dataset settings}
\resizebox{\columnwidth}{!}{%
\begin{tabular}{@{}cccc@{}}
\toprule
Settings & Parameters & SMPL-NPT & SMG-3D \\ \midrule
\multirow{3}{*}{Training} & Identity & 1-16 & 1-35 \\
 & Pose & 1-400 & 1-180 \\
 & Used pairs & 8000 & 4000 \\ \midrule
\multirow{3}{*}{Testing seen} & Identity & 17-30 & 36-40 \\
 & Pose & 1-400 & 1-180 \\
 & Used pairs & 72 & 100 \\ \midrule
\multirow{3}{*}{Testing unseen} & Identity & 17-30 & 36-40 \\
 & Pose & 401-600 & 181-200 \\
 & Used pairs & 72 & 100 \\ \bottomrule
\end{tabular}
}

\end{table}

\textbf{Dataset Settings.} The details of sample settings are given in Table \ref{Tab:dataset settings}. For the SMPL-NPT dataset, we use the same protocol as \cite{NPT} where 16 identities and 400 poses are used for training. Since it results in more than 40,000,000 potential training pairs which is way larger than our computational capacity, we randomly select 8,000 training pairs at each epoch during the training. Similarly, it becomes 4,000 pairs for the SMG-3D dataset. There are 72 pairs and 100 pairs used to test the results on these two datasets, the pose and identities used for the ``seen'' and ``unseen'' settings in the testing phase can be found in Table \ref{Tab:dataset settings}.

\begin{table*}[tbp]
\caption{Training settings for all the datasets. ``GCT'', ``LIR'', and ``LR'' stand for the GC-Transformer, latent isometric regularization module, and learning rate, respectively.}
\label{Tab:training settings}
\centering
\begin{tabular}{@{}ccccccccc@{}}
\toprule
Parameters & \multicolumn{2}{c}{SMPL-NPT} & \multicolumn{2}{c}{SMG-3D} & \multicolumn{2}{c}{SMAL} & \multicolumn{2}{c}{FAUST \& MG-cloth} \\ \midrule
 & GCT & LIR & GCT & LIR & GCT & LIR & GCT & LIR \\
Epochs & 1000 & 400 & 1000 & 400 & 200 & 100 & - & 200 \\
LR & $5\times 10^{-5}$ & $5\times 10^{-5}$ & $5\times 10^{-5}$ & $5\times 10^{-5}$ & $5\times 10^{-5}$ & $5\times 10^{-5}$ & - & $5\times 10^{-5}$ \\
LR milestones & {[}200,500{]} & - & {[}400,600{]} & - & - & - & - & - \\
LR decay & 0.1 & - & 0.1 & - & - & - & - & - \\
$\lambda_{constra}$ & $5\times 10^{-4}$ & $5\times 10^{-3}$ & $5\times 10^{-4}$ & $5\times 10^{-3}$ & $5\times 10^{-4}$ & $5\times 10^{-3}$ & - & $5\times 10^{-3}$ \\
Batch size & 8 & 8 & 8 & 8 & 8 & 8 & - & 8 \\ \bottomrule
\end{tabular}

\end{table*}

\textbf{Hyper-Parameters.} 
Our algorithm is implemented in PyTorch~\cite{paszke2019pytorch}. All the experiments are carried out on a PC with a single NVIDIA Tesla V100, 32GB. We train our networks for 1000 epochs with a learning rate of 0.00005 and Adam optimizer. The weight settings in the paper are  $\lambda_{rec}{=}1$, $\lambda_{edge}{=}0.0005$, and $\lambda_{contra}{=}0.0005$.
The batch size is fixed as 8 for all the settings. Training time is around 100-150 hours. The hyper-parameters for training on all datasets are presented in Table \ref{Tab:training settings}. Since we directly evaluate the model on the FAUST and MG-cloth datasets without training, only LIR is trained for the latent pose space normalization within the datasets. Note that, batch size of 8 is only available with 32GB memory GPUs to run the GC-Transformer. For GPUs with 12 or 24GB memory, the batch size should be adjusted to 2-6 according to our experience.

\subsection{Time Consumption}
Theoretically, it will take 300 hours to train our GC-Transformer for 1,000 epochs on a single GPU of NVIDIA Tesla V100, 32GB (with 8,000 samples). However, ``early stop'' is possible around 200-400 epochs (70-120 hours) with which the PMD values are already equal to or even better than the results reported in the paper (e.g., 0.6 and 4.0 for the SMPL-NPT dataset). Further training will consistently gain but little improvement of the performance (lower PMD values). Thus, in practice we can finish the training of the models within 70-120 hours. The training time for LIR module (200 epochs with 400 samples) is around 2 hours.

\section{SMG-3D Dataset}
\label{sec:2}
Here we give more details and examples of our SMG-3D dataset.

\subsection{Dataset Introduction}

\begin{figure*}[!h]
    \centering
    \includegraphics[width=1\linewidth]{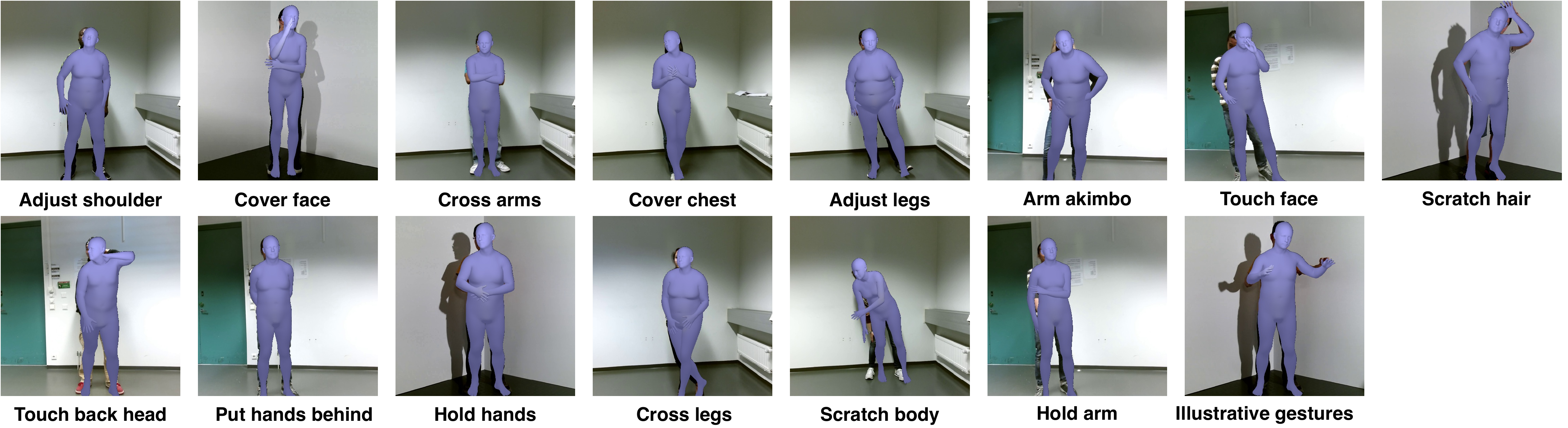}
    \caption{15 body gesture classes used for generating 3D meshes of our SMG-3D dataset. We generate 3D body meshes from 2D images with STRAPS \cite{STRAPS} and enhanced by \cite{SMPLX}. 200 different poses from those 15 classes performed by 40 participants are used as template pose meshes in our SMG-3D dataset.}
    \label{fig:SMG3d}
\end{figure*}

Our SMG-3D dataset is derived from an existing 2D body pose dataset, i.e., SMG~\cite{SMG}. SMG consists of 3,699 gesture clips with 17 classes performed during daily communication. We transformed SMG into a 3D body mesh dataset SMG-3D in a semi-synthesized way. We first estimate preliminary 3D meshes with the model STRAPS~\cite{STRAPS} from all the 3,699 original 2D images of 17 classes from SMG as shown in Fig~\ref{fig:SMG3d}. Then, we select 200 poses (including 15 classes) as template poses to form the potential pose space and enhanced them by Vposer~\cite{SMPLX}. Vposer~\cite{SMPLX} is a human body prior learning model that can generate valid and natural 3D human poses based on the human body prior. This body prior can be obtained by training Vposer on the existing datasets like those 3,699 3D meshes from SMG or other larger dataset like AMASS \cite{AMASS}. Next, we decompose the generated 3D meshes into 40 distinct identity registrations and 200 pose registrations as latent parameters that can be applied to SMPL model~\cite{SMPL}. By pairing the identities and poses in the latent space, 8,000 body meshes are synthesized via SMPL model~\cite{SMPL} as shown in Fig.~\ref{fig:identitesandposes}. 

\subsection{Identity and Pose Registrations}

\begin{figure*}[!h]
    \centering
    \includegraphics[width=1\linewidth]{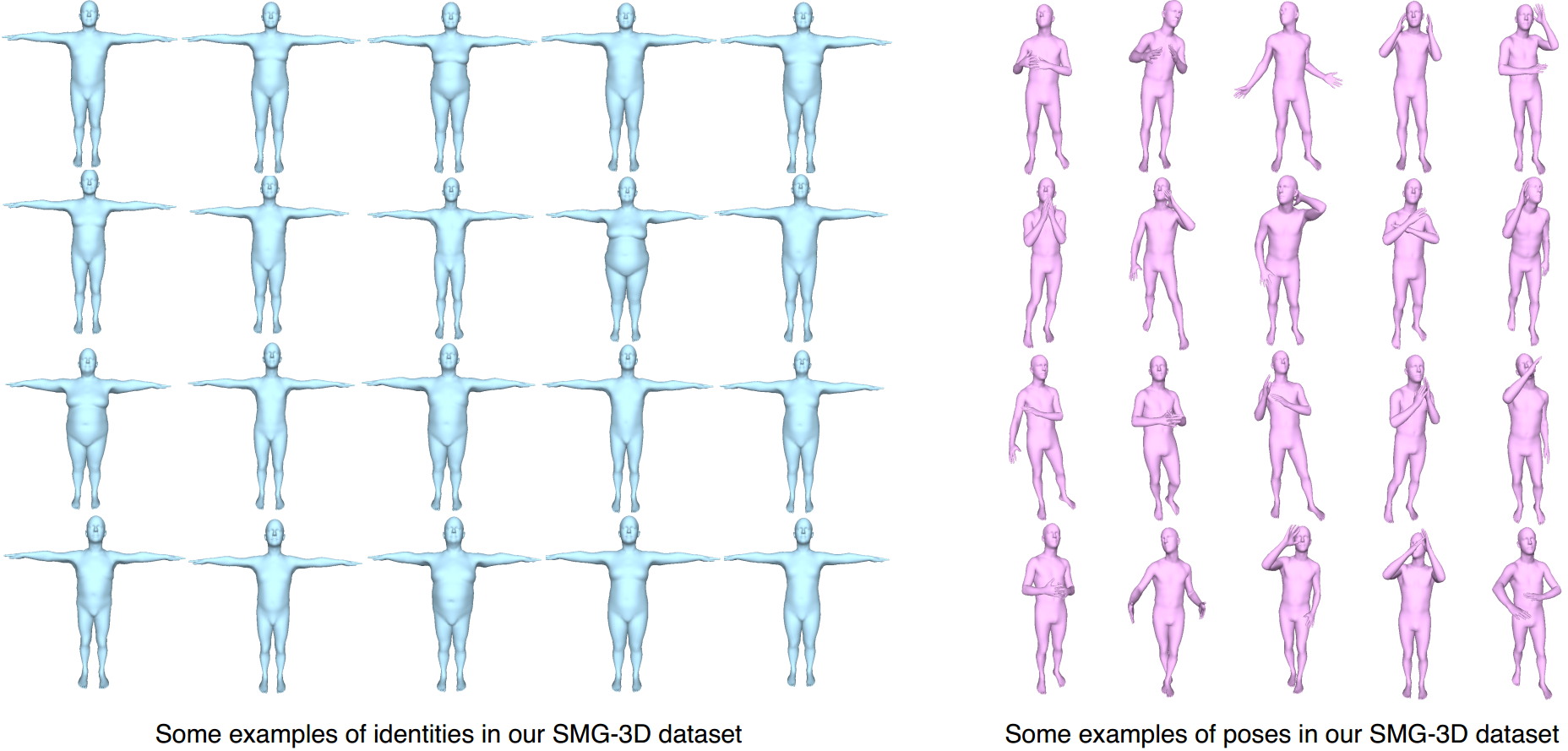}
    \caption{The samples identities and poses used as latent space for generating 3D meshes of our SMG-3D dataset. Only first 20 identities and 20 poses are presented in this figure. There are 40 identities and 200 poses used in practical work.}
    \label{fig:identitesandposes}
\end{figure*}

40 distinct identity registrations and 200 different pose registrations are decomposed from 2D image estimations. The visualization of some of the identities and poses are shown in Fig.~\ref{fig:identitesandposes}, in which all the identities are made into standard poses ($\beta {=} 0$) and poses made into a unified identity. There are $10$ coefficients to define one identity registration and $24{\times}3$ coefficients to define one pose registration. 


\section{Additional Pose Transfer Results}
\label{sec:3}
Here we present more pose transfer results. First, additional examples of transferring poses from existing SMPL-NPT dataset to unseen datasets are given. Then, we show some extra examples on SMAL animal datasets. At last, the performance of our method with other state-of-the-art methods and ground truth on the SMPL-NPT dataset. Note that, for convenience, all the models used below are all stacked with max-pooling layers for adapting meshes with various vertex numbers.

\subsection{Pose Transfer to Unseen Datasets}

\begin{figure*}[!h]
    \centering
    \includegraphics[width=0.85\linewidth]{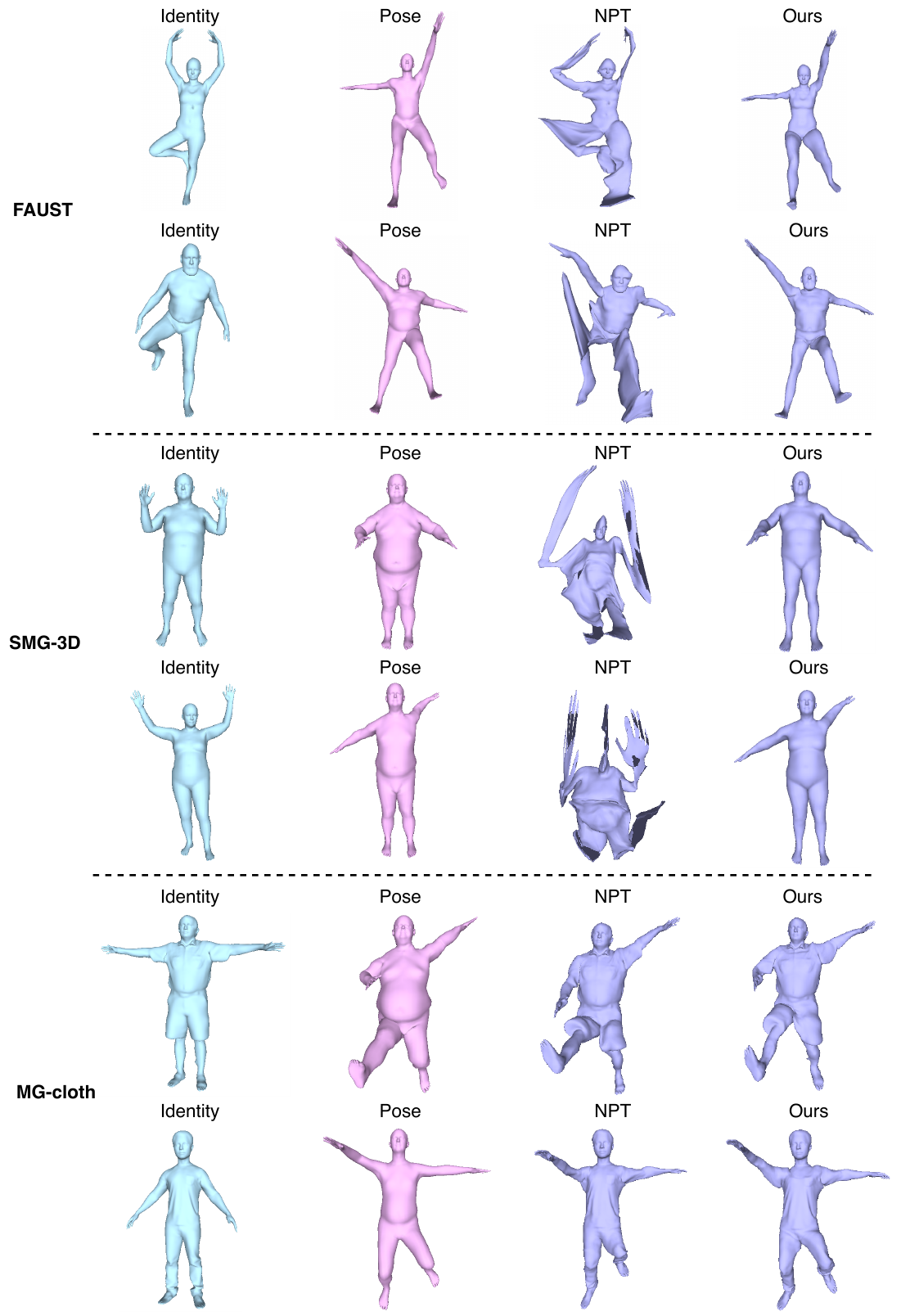}
    \caption{Additional examples of cross-dataset pose transfer. The models are directly evaluated on unseen datasets including FAUST \cite{FAUST}, SMG-3D, and MG-cloth \cite{MG-cloth}.}
    \label{fig:unseen}
\end{figure*}

Here we present more qualitative examples of pose transfer results on unseen datasets in Fig.~\ref{fig:unseen}. For all the three datasets, our performances has significantly better visual effects than the leading method NPT~\cite{NPT}, which further validates the effectiveness of the proposed method.

\subsection{Generalized Pose Transfer on SMAL}

\begin{figure*}[!h]
    \centering
    \includegraphics[width=1\linewidth]{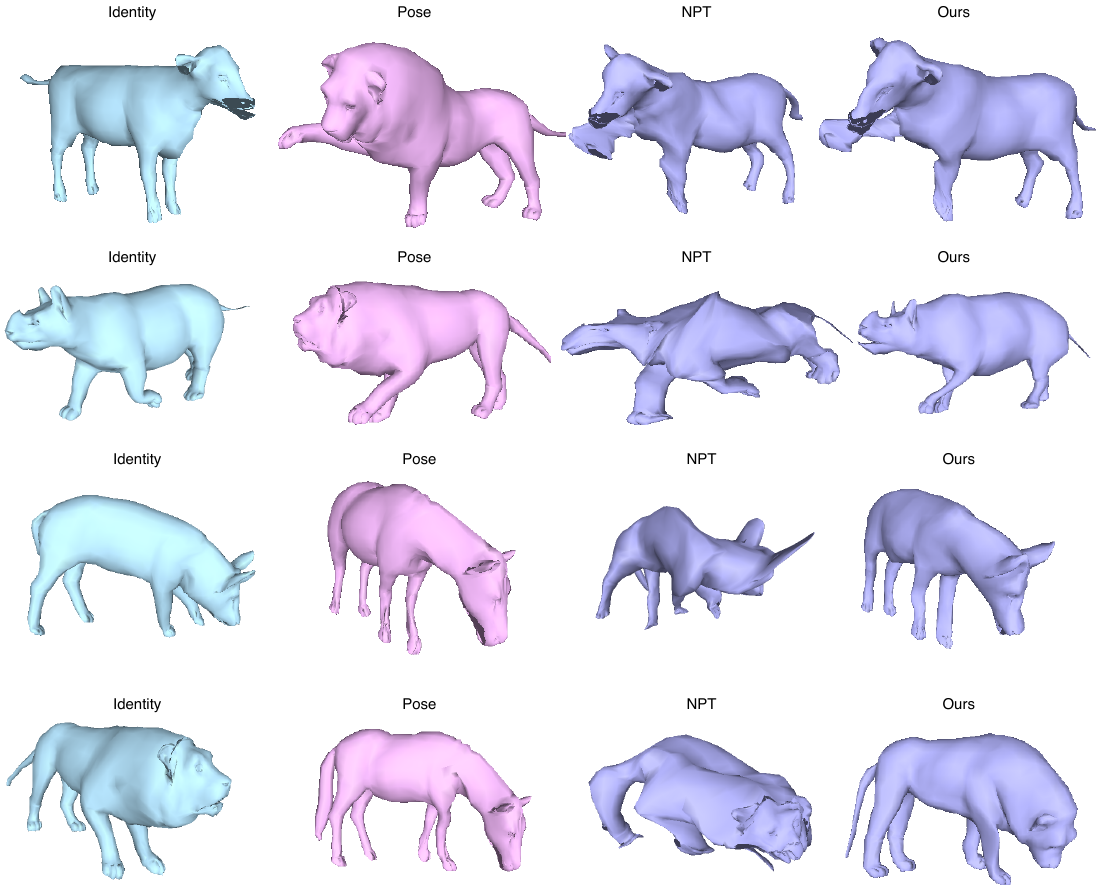}
    \caption{Additional examples of non-human pose transfer. We carry out the experiments on the SMAL \cite{SMAL} animal dataset. Extra training for all the models are conducted to adapt the domain.}
    \label{fig:smal}
\end{figure*}

Here we present more examples of pose transfer results on the animal dataset SMAL \cite{SMAL}. As shown in Fig.~\ref{fig:smal}, our method can achieve robust pose transfer even in different domains. It is evident that our model can performance robust pose transfer even on different domains.

\begin{figure*}[!h]
    \centering
    \includegraphics[width=1\linewidth]{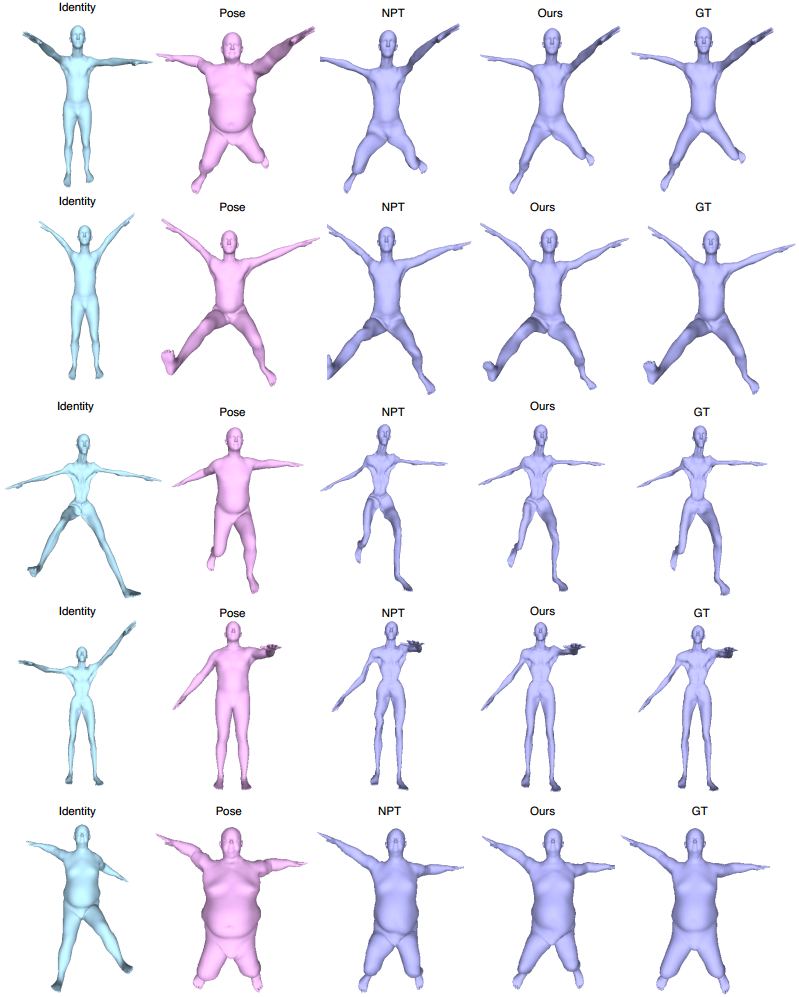}
    \caption{Additional examples of ``seen poses'' on the SMPL-NPT dataset.}
    \label{fig:SMPLseenpose}
\end{figure*}

\begin{figure*}[!h]
    \centering
    \includegraphics[width=1\linewidth]{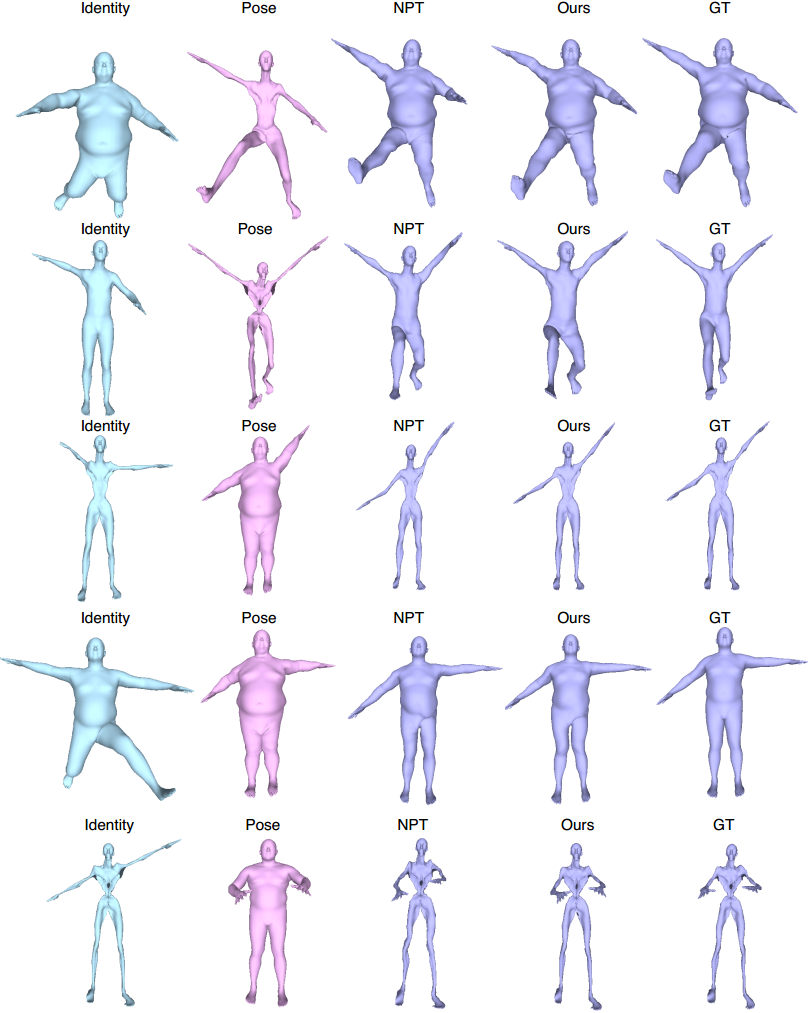}
    \caption{Additional examples of ``unseen poses'' on the SMPL-NPT dataset.}
    \label{fig:SMPLunseenpose}
\end{figure*}

Lastly, we compare our method with the start-of-the-art NPT model \cite{NPT} as well as ground truth in Fig.~\ref{fig:SMPLseenpose} and Fig.~\ref{fig:SMPLunseenpose}. Fig.~\ref{fig:SMPLseenpose} shows the performance on ``seen pose'' setting that the desired pose is available in the training set. Fig.~\ref{fig:SMPLunseenpose} shows the performance on ``unseen pose'' setting that the desired pose is not available in the training set. For both settings, our performances are closer to the ground truths.

\end{document}